\documentclass[10pt]{article}
\usepackage{amsmath}
\usepackage{amsfonts}
\usepackage{amssymb}
\usepackage{mathrsfs}
\usepackage{booktabs} 
\usepackage{array}    

\usepackage{siunitx}  
\usepackage{amsthm}
\usepackage{MnSymbol}
\newtheorem{theorem}{Theorem}[section]

\usepackage[cal=boondox,scr=boondoxo]{mathalfa}
\usepackage{amsmath, amssymb}
\usepackage{float}
\usepackage{subfig}
\usepackage{fancybox,graphicx}
\usepackage{subfig}
\usepackage{caption}
\usepackage{color}
\usepackage{authblk}

\usepackage{algorithm,algpseudocode}

\usepackage[colorlinks]{hyperref}

\usepackage{accents}
\usepackage[titletoc,title]{appendix}
\usepackage{cite}
\usepackage[english]{babel} 
\usepackage[top=2in, bottom=1.5in, left=1in, right=1in]{geometry}

\usepackage{physics}  
\usepackage{subcaption}
\title{Revisiting Reset Mechanisms in Spiking Neural Networks for Sequential Modeling:  Specialized Discretization for Binary Activated RNN} 
 
\author{
    Enqi Zhang \\
   \small University of Electronic Science and Technology of China\\
   \small School of Computer Science and Engineering \\
    \small \texttt{enqizzz@std.uestc.edu.cn }
}

\begin{document}
\date{}  

    \maketitle
\begin{abstract}
In the field of image recognition, spiking neural networks (SNNs) have achieved performance comparable to conventional artificial neural networks (ANNs). In such applications, SNNs essentially function as traditional neural networks with quantized activation values. This article focuses on an another alternative perspective,viewing SNNs as binary-activated recurrent neural networks (RNNs) for sequential modeling tasks.From this viewpoint, current SNN architectures face several fundamental challenges in sequence modeling: (1) Traditional models lack effective memory mechanisms for long-range sequence modeling; (2) The biological-inspired components in SNNs (such as reset mechanisms and refractory period applications) remain theoretically under-explored for sequence tasks; (3) The RNN-like computational paradigm in SNNs prevents parallel training across different timesteps.To address these challenges, this study conducts a systematic analysis of the fundamental mechanisms underlying reset operations and refractory periods in binary-activated RNN-based SNN sequence models. We  re-examine whether such biological mechanisms are strictly necessary for generating sparse spiking patterns, provide new theoretical explanations and insights, and ultimately propose the fixed-refractory-period SNN architecture for sequence modeling.
\end{abstract}

\section{Introduction}

The application of spiking neural networks (SNNs) in artificial intelligence has been maturing with increasing research attention. Currently, SNN models have achieved performance comparable to traditional artificial neural networks (ANNs) in static tasks such as image classification \cite{wu2018spatio,yao2023spike} and audio signal processing\cite{wang2025ternary,zhang2024spike}. Focusing on AI applications rather than brain-like intelligence research, we can categorize SNNs into two distinct perspectives.

The first perspective views SNNs as traditional neural networks with quantized activations. For tasks like image classification, this approach introduces an additional virtual temporal dimension, using firing rates across multiple timesteps as information carriers. Since the information is transmitted through discrete frequency values, these models exhibit similar properties to quantized activation convolutional neural networks.

The second perspective, which is the focus of this study, considers SNNs as binary-activated recurrent neural networks (RNNs). Here, there's no artificial temporal dimension - only the inherent temporal dimension of the input sequence. The spiking output contains not just frequency information but precise temporal information. From this viewpoint, four critical questions emerge:

1.How to understand the sequential memory capacity of binary-activated RNN-style SNNs

2.The fundamental nature of information transmission between SNN layers via spiking sequences

3.The role of reset mechanisms and refractory periods in SNNs

4.Overcoming the limitation of non-parallelizable training across timesteps

Regarding the first question, existing work on state space models\cite{gu2020hippo,gu2021efficiently,orvieto2023resurrecting} and traditional mathematical tools (Fourier transforms, wavelet analysis, etc.) have provided answers, we will provide a brief explanation in the appendix.~\ref{sec:memory} based on these studies. This allows us to decouple SNNs into two independent modules: a sequence memory module and a spiking module, offering new insights into SNN sequence modeling.

For the second question, which concerns how to understand the mapping of arbitrary input spike sequences by spiking neural networks.Take frequency-encoded SNNs for static images as an example, in essence, such networks can be understood as mapping discrete frequency values to another set of discrete frequency values. However, for sequences containing temporal information, what approach should be adopted to map such sequences? Should we use the Leaky Integrate-and-Fire (LIF) model, the Integrate-and-Fire (IF) model, or another method? Moreover, how could we understand the mapping between sequences? This paper approaches the problem from the perspective of probability distributions, treating the information to be transmitted as a distribution and spikes as discrete sampling points of this distribution. In this way, we can interpret the transmission of information between layers.Existing research has employed methods where input distributions are sampled to generate spikes as firing events\cite{bal2024p}. The difference between this work and ours is that our study treats spikes merely as sampling points, serving as an approximation of the input and functioning as an analytical tool. The goal of our research is to construct a deterministic system, unlike the non-deterministic system proposed in \cite{bal2024p}, which incorporates stochastic elements.

The third and fourth questions concern how reset mechanisms and refractory periods hinder parallel training. We interpret these mechanisms as performing additional sparse sampling on discrete sampling points to achieve sparsity. This perspective leads to our fixed-refractory-period SNN model and its improved variant, spikingPssm, which maintains sparsity while enabling parallel training.

Our final spikingPssm model essentially combines state space models with a specialized PSN module \cite{fang2023parallel}. While architecturally not revolutionary, it achieves competitive results on sequential CIFAR-10 (L=1024). Though not state-of-the-art, our goal is to reframe understanding of SNN sequence modeling. The success of this simple architecture,using basic linear dynamical systems for memory and PSN for spiking, raises questions about whether complex nonlinear dynamics or opaque spiking mechanisms are truly necessary. Notably, high-performing SNNs in AI applications largely emulate ANNs (either as quantized ANNs or binary RNNs), suggesting the fundamental nature of spikes warrants further investigation. We hope this work provides valuable insights for the community.

\section{Related Works}

Spiking neural networks have demonstrated strong performance in static input tasks or scenarios requiring no long-term memory retention. However, their development remains limited in sequential tasks demanding long-range dependency modeling. One effective approach has been the incorporation of attention mechanisms\cite{vaswani2017attention}. In traditional artificial neural networks, attention mechanisms address vanishing gradients and long term memory limitations in recurrent architectures by establishing direct point-to-point relationships, achieving remarkable success across tasks despite quadratic computational complexity. Several SNN studies have similarly adopted attention to enhance long-term memory capacity \cite{qin2023attention}.Nevertheless, conventional attention-based models fail to compress historical information efficiently, incurring substantial computational overhead. Moreover, these SNN implementations often underutilize intrinsic neuronal dynamics\textemdash the leaky integrate-and-fire (LIF) model primarily functions as a quantized encoder, exhibiting learning principles scarcely distinct from traditional ANN-based Transformers. This raises a critical question: How can SNNs leverage their inherent neuronal dynamics to achieve long-term memory with low storage complexity? Previous work has explored adaptive spiking neurons \cite{yin2021accurate,bellec2018long,shaban2021adaptive,zhang2025spiking}through learnable time constants or threshold optimization, yet these efforts lack systematic analysis from an information storage perspective.

The emergence of Mamba~\cite{gu2023mamba} in 2024 has revitalized interest in state space models (SSMs)~\cite{gu2021combining,gu2021efficiently}, offering novel insights into sequential memory. Beginning with the HiPPO theory~\cite{gu2020hippo}, which formalized real-time sequence memory updates via orthogonal polynomial projections, subsequent integration with SSMs yielded progressively simplified architectures excelling in long-sequence tasks. Mamba's selective mechanism further enabled recurrent neural networks to mimic attention-like functionality. These models~\cite{orvieto2023resurrecting,smith2022simplified,qin2023hierarchically} leverage the parallelizability of linear dynamical systems during training while maintaining $\mathcal{O}(L)$ inference complexity, presenting a viable alternative to attention. Similarly, linear attention variants~\cite{katharopoulos2020transformers,shen2021efficient,ma2022mega,yang2024parallelizing} compress historical information into fixed-memory representations, contrasting with traditional attention's unconstrained memory growth, reinvigorating research into memory mechanisms.

Advancements in artificial neural networks have directly inspired spiking neural network (SNN) architectures. Mirroring state space models (SSMs), bio-inspired multi-compartmental or dendritic structures (e.g., TC-LIF~\cite{zhang2024tc}, PMSN~\cite{chen2024pmsn}, DH-LIF~\cite{zheng2024temporal}) have spawned attention-free, constant-memory SNN models. Notably, PMSN employs an integrate-fire (IF) output layer to resolve SNNs' parallel training limitations. Following SSM frameworks, Stan~et~al.~\cite{stan2024learning} eliminated reset and refractory mechanisms via binary activations, while P-SpikeSSM~\cite{bal2024p} introduced stochastic spiking into SSMs, establishing a new paradigm. SpikeSSM~\cite{zhong2024spike} and SpikingSSM~\cite{shen2024spikingssms} achieved parallel training and sparse spiking through learned firing functions and max-min boundary compression, respectively.
Despite these advances, fundamental questions persist:
What is the functional role of spikes in SNNs from an applied AI perspective (rather than a neuro-mimetic one)?
For sequence modeling with native temporal dimensions (unlike static image tasks with artificial timesteps), what information do spike signals convey between layers?
How should reset mechanisms and refractory periods be interpreted in practical applications?
The application of spiking neural networks in practical tasks may not be limited to merely simulating biological systems. Providing theoretical definitions for their operation is what truly bridges computational neuroscience with artificial intelligence applications.
\section{Background}
\subsection{LIF model}
The Leaky Integrate-and-Fire (LIF) model, widely adopted in computational neuroscience and spiking neural networks, achieves an optimal balance between computational efficiency and biological plausibility by abstracting the complex ion channel dynamics of the Hodgkin-Huxley model into basic electrical components. The core differential equation of the model is:
\begin{equation}
\tau_m \frac{d V(t)}{d t} = -V(t) +  I(t)
\end{equation}
Here, 
$V$ represents the membrane voltage, 
$I$ denotes the input current, and $\tau_m$ is a time constant controlling the decay rate of the membrane voltage when no input is present.For numerical solution, discretization is typically employed:
\begin{equation}
V(t + \Delta t) = V(t) + \frac{\Delta t}{\tau_m}\left(-V(t) + I(t)\right)
\end{equation}
where $\Delta t$ represents the simulation timestep. When the membrane potential $V(t)$ exceeds the threshold potential $V_{\text{th}}$, the neuron emits a discrete spike output $\delta(t - t_{\text{spike}})$ and resets according to:
\begin{equation}
    V(t_{\text{spike}}^+) = V_{\text{reset}} 
    \label{hardreset}
\end{equation}
\begin{equation}
    V(t_{\text{spike}}^+) = V(t_{\text{spike}}^-) - V_{th}
    \label{softrest}
\end{equation}
Equation~(\ref{hardreset}) represents the \textbf{hard reset} mechanism where the membrane potential is directly reset to the resting potential, while Equation~(\ref{softrest}) describes the \textbf{soft reset} approach that subtracts the threshold potential $V_{\text{th}}$ from the current membrane potential upon firing.
Currently, the most widely adopted spiking neuron model in neural networks is the Leaky Integrate-and-Fire model. The complete network dynamics can be formally expressed as \cite{fang2021incorporating}:
\begin{equation}
    \begin{aligned}
    U_i^l(t+1) &= (1- \frac{1}{\tau} )V_i^l(t) + \sum_j w_{ij}^l s_j^{l-1}(t+1)  \\
    s_i^l(t+1) &= \Theta(U_i^l(t+1)-V_{th}) \\
    V_i^l(t+1) &= (1-s_i^l(t+1))U_i^l(t+1) + s_i^l(t+1)V_{reset} 
    \label{snn_net}
\end{aligned}
\end{equation}
Equation~(\ref{snn_net}) describes the current mainstream SNN computation approach, sequentially representing: (1) the membrane potential dynamics after weighted summation of input currents, (2) the threshold comparison using the step function, and (3) the hard reset mechanism after spiking. Here, both $U[t]$ and $V[t]$ represent membrane potential, $w_{ij}$ denotes the synaptic weight from neuron $j$ in the previous layer to neuron $i$ in the current layer, $s_i[t]$ indicates the spike output at time $t$, and $V_{\text{reset}}$ represents the reset potential. This describes the hard reset case. The $\Theta(\cdot)$ function represents the Heaviside step function, where if $U_i^l[t+1]$ exceeds the threshold $V_{\text{th}}$, it outputs 1; otherwise the model remains silent with 0 output.
\subsection{Two Perspectives of Spiking Neural Networks}

\textbf{The Perspective of activation-quantized ANN}

\begin{figure}[H]
    \centering
    \includegraphics[width=0.75\linewidth]{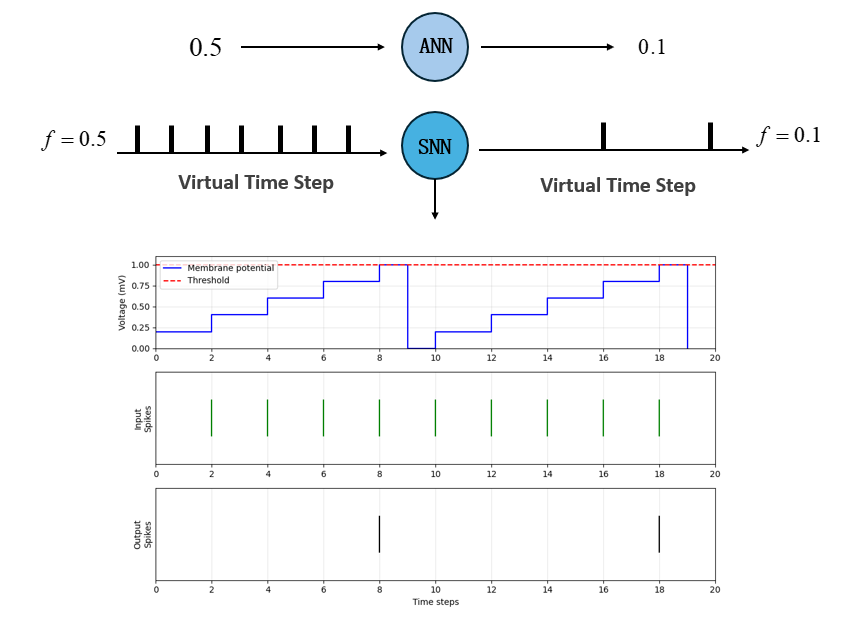}
     \caption{The Perspective of activation-quantized ANN}
    \label{fig:qasnn}
\end{figure}

From this perspective, the temporal dimension in SNNs can be regarded as a \textit{virtual time axis}. By introducing this additional virtual time dimension, SNNs can approximate the floating-point activations in traditional neural networks through firing rates. This approach enables SNNs to achieve excellent performance in static image processing tasks, though their functionality differs fundamentally from recurrent neural networks for sequential data processing.
As illustrated in Fig.~\ref{fig:qasnn}, rate-coded SNNs exhibit remarkable similarity to ANNs in their fundamental characteristics. The input and output values of ANNs are constrained to discrete levels $\{0.0, 0.1, 0.2, \dots, 0.9, 1.0\}$, which corresponds closely to the frequency-domain representation of SNNs with limited timesteps. For instance, an SNN converting an input spike train with 0.5 firing rate to an output with 0.1 firing rate is functionally equivalent to an ANN mapping input 0.5 to output 0.1. This conceptual parallel has inspired several ANN-SNN co-training algorithms \cite{meng2022training,wu2021tandem}.

In such implementations, the spiking architecture of SNNs essentially functions as a \textit{quantization mechanism} rather than a dynamic system for temporal sequence processing. Furthermore, time-to-first-spike (TTFS) encoded SNNs can also be viewed as equivalent to quantized activation ANNs, particularly for two-stage TTFS-based models\cite{stanojevic2024high,wei2023temporal,yang2023lc,zhang2021rectified}. These models fundamentally operate as \textit{declarative networks} \cite{gould2021deep}, as detailed in the Appendix.~\ref{sec:ttfs}, demonstrating their equivalence to quantized activation declarative networks. 

This analysis reveals that in static image processing, regardless of temporal or rate coding schemes, spiking neural networks maintain intrinsic connections with quantized activation ANNs.

\textbf{The Perspective of binary-activated RNN}

A system that evolves over time is termed a \emph{dynamical system}, typically described by differential equations. The development of neural networks has maintained close connections with advances in dynamical systems theory. The \emph{Neural Ordinary Differential Equation} (Neural ODE) \cite{chen2018neural,dupont2019augmented} represents a deep learning framework that models continuous dynamical systems through ordinary differential equations. Its fundamental innovation lies in replacing the discrete layer structure of traditional neural networks with continuous dynamics,employing ODE solvers for both forward propagation and backpropagation, or using ODE to learn lessons from nonlinear dynamic systems.

Given input data $\mathbf{x}(t_0) \in \mathbb{R}^d$, a Neural ODE is formally defined as:
\begin{equation}
\frac{d\mathbf{x}(t)}{dt} = f_\theta(\mathbf{x}(t)), \quad t \in [t_0, t_1]
\label{eq:node}
\end{equation}
where $f_\theta: \mathbb{R}^d \to \mathbb{R}^d$ represents a neural network-parameterized vector field,$\mathbf{x}(t_1)$ is obtained through numerical integration:
    \begin{equation}
        \mathbf{x}(t_1) = \mathbf{x}(t_0) + \int_{t_0}^{t_1} f_\theta(\mathbf{x}(t)) dt
    \end{equation}
Regarding this differential equation, recurrent neural networks also represent time-varying systems, and thus their intrinsic properties can be analyzed through this differential equation framework. When interpreting network layers as temporal dimensions, residual neural networks (ResNets)\cite{he2016deep} can similarly be viewed as dynamical systems \cite{chen2018neural}. Different differential equations and discretization strategies yield distinct network architectures \cite{yi2023nmode,sander2021momentum}.

This work is closely related to state space models, which constitute linear dynamical systems. The governing equations of linear dynamical systems can be expressed as:
\begin{equation}
\frac{dh(t)}{dt} = Ah(t) + Bu(t)
\label{eq:cont}
\end{equation}
$h(t) \in \mathbb{R}^n$ is the state vector and $u(t) \in \mathbb{R}^m$ is the input. The linear dynamical system is the simplest form of dynamical system.

\begin{figure}[H]
    \centering
    \includegraphics[width=0.5\linewidth]{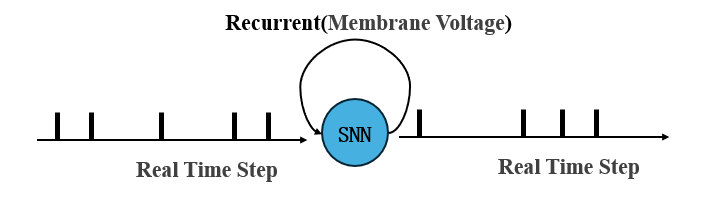}
    \caption{The Perspective of binary-activated RNN}
    \label{fig:rsnn}
\end{figure}

Clearly, when disregarding reset mechanisms, the simplest LIF model in SNNs constitutes a linear dynamical system. To fully utilize the dynamical characteristics within SNNs,differing from viewing SNNs as quantized ANNs,we consider SNNs as binary-activated recurrent neural networks and analyze them from the perspective of dynamical systems. This approach allows better understanding of SNNs' potential in processing temporal sequence tasks, particularly for capturing temporal dynamic features.
As shown in Fig.~\ref{fig:rsnn}, the spike generation process in SNNs can be viewed as a binarized RNN dynamical system. In such a system, each neuron's spiking behavior depends not only on current inputs but also on its internal state and past spiking history. Note that the time axis here represents real time, not virtual time. The input and output sequences contain not only frequency information but also rich temporal information. The Recurrent module refers to the dynamical behavior within SNNs, rather than feedback connections of spike sequences \cite{xiao2021training}. This can be either a simple linear dynamical system or other complex nonlinear dynamical systems. Such dynamical characteristics may give SNNs advantages in processing temporal sequence tasks, especially those requiring capture of complex temporal dependencies.

\subsection{State Space Models}

State Space Models (SSMs) are mathematical frameworks for describing dynamical systems. In recent years, SSMs \cite{gu2021efficiently} have been introduced into deep learning as a novel neural network architecture particularly suitable for sequential data processing. The continuous-time formulation is given by:

\begin{equation}
    \frac{dh(t)}{dt} = A h(t) + B x(t)
\end{equation}
\begin{equation}
    y(t) = C h(t) + D x(t)
    \label{output_ssm}
\end{equation}

where $x$ represents a one-dimensional continuous input sequence, $t$ denotes the time dimension, with $A \in \mathbb{R}^{N \times N}$, $B \in \mathbb{R}^{N \times 1}$, and $C \in \mathbb{R}^{1 \times N}$. In practice, the model typically processes discrete one-dimensional sequences $\{x_1,x_2,...,x_n\}$, transforming them into one-dimension output sequences $\{y_1,y_2,...,y_n\}$.

For computational efficiency, this paper follows the mainstream approach of state space models \cite{gu2021efficiently,gu2023mamba} by adopting the zero-order hold (ZOH) discretization method, resulting in the following discrete-time state space representation:

\begin{align}
    h(t) &= \hat{A} h(t-1) + \hat{B} x(t)\\
    \hat{A} &= e^{\Delta A}\\
    \hat{B} &= A^{-1}(e^{\Delta A} - I)B
    \label{AB}
\end{align}

Here $\Delta$ represents the discretization step size. The output equation (\ref{output_ssm}) remains unchanged. The $Dx(t)$ term can be omitted as it can be replaced by residual connections.

During training, SSMs enable parallel processing of all timesteps' inputs, avoiding the sequential computation of traditional RNNs and backpropagation through time (BPTT), significantly accelerating training. With initial state $h(0) = \mathbf{0}$, the output at any timestep can be expressed as:

\begin{equation}
    h(t) = \sum_{k=0}^{t} \hat{A}^{t-k} \hat{B} x_k
\end{equation}

Since each output depends only on previous inputs (not outputs), the layer can be computed as a convolution with kernel:

\begin{equation}
\hat{K} = \left( C\hat{B}, C\hat{A}\hat{B}, \ldots, C\hat{A}^{L-1}\hat{B} \right) \in \mathbb{R}^{1 \times L}
\end{equation}

yielding the output:

\begin{equation}
    y(k) = \sum_{j=0}^k \hat{K}(j) x(k-j)
\end{equation}

The subsequent Gated Linear Unit (GLU) operations, being non-recurrent, maintain this parallelizability. While the naive computation requires $O(L^2)$ multiplications, the special structure allows acceleration via Fast Fourier Transform (FFT), reducing complexity to $O(L\log L)$ \cite{gupta2022diagonal}.

\section{Theoretical Analysis}
\subsection{Decoupling Spiking and Memory Modules in Spiking Neural Networks}

To understand the essential mechanisms of information storage and transmission in spike trains, as well as the nature of reset operations in artificial intelligence applications, we must first address a fundamental question: how exactly does a spiking neural network transform an input spike train into an output spike train? Consider the simplest case of linear transformation - mapping a vector in Euclidean space to another through scaling or rotation. Traditional artificial neural networks perform similar vector mappings in Euclidean space, albeit nonlinearly: first a linear transformation followed by a nonlinear activation function. Similarly, rate-coded SNNs processing static datasets can be viewed as mapping vectors whose elements represent firing frequencies, making them functionally analogous to conventional ANNs. However, SNNs directly processing temporal sequences differ significantly. Their input spike trains contain not just frequency information but rich temporal patterns. This chapter provides a detailed analysis of how binary-activated RNN-style SNNs perform such spike-to-spike mappings for arbitrary input sequences.

Before formal analysis, we establish a key definition: \textbf{For any recurrent neural network processing sequential data, we consider the entire input sequence as a \emph{distribution}, where each recurrent layer transforms this distribution into another distribution.}
This perspective allows us to analyze RNNs holistically, analogous to how CNNs or fully-connected networks process static images. While distribution values nominally range between 0 and 1, post-activation values may exceed 1 (treated as 1 here). Note this probabilistic interpretation serves only as an analytical tool, not as a basis for building actual stochastic spiking models.Therefore, a rigorous definition is unnecessary here,this serves merely as an interpretive framework.

\begin{figure}[H]
    \centering
    \includegraphics[width=0.75\linewidth]{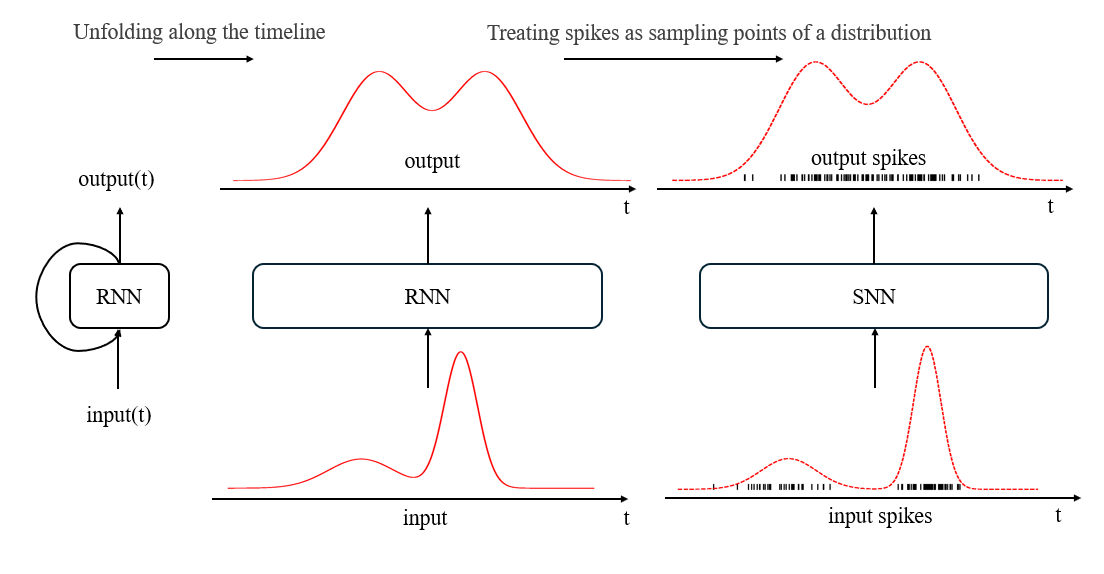}
    \caption{Modeling Recurrent Architecture Inputs as Probability Distributions}
     \label{fig:distributionrnn}
\end{figure}

As shown in Fig.~\ref{fig:distributionrnn}, from the perspective of traditional recurrent neural networks, at each timestep the network receives input and performs a nonlinear transformation on the weighted combination of historical information and current input. This makes it difficult to conceptualize long sequence modeling as processing either a very long image or the extended distribution illustrated in Fig.~\ref{fig:distributionrnn}. However, recently prominent linear RNN models \cite{orvieto2023resurrecting,gu2021efficiently,orvieto2023universality}(including State Space Models) have addressed this issue.
These models first encode long sequences through a linear dynamical system. Due to the parallel inference capability of linear systems, they can simultaneously obtain outputs for any timestep. After processing through the linear dynamical system layer, they apply a pointwise nonlinear transformation such as MLP or Gated Linear Unit (GLU) \cite{dauphin2017language}, thereby achieving nonlinear transformation for the entire system. From the linear RNN perspective, it becomes straightforward to establish correspondence between long sequences and static input images. Numerous works\cite{orvieto2023universality,wang2023state}have demonstrated the approximation capabilities of this RNN formulation.

Recent studies have explored generating spikes by sampling input distributions\cite{bal2024p}. The difference from our work lies in that our research merely treats spikes as sampling points for input approximation, serving as an analytical tool. Our objective is to construct a deterministic system rather than the stochastic system incorporating random factors as in \cite{bal2024p}. 

This viewpoint offers valuable insights into the memory mechanisms of spiking neural networks. The memory capacity of linear systems has been thoroughly investigated in prior work \cite{gu2021efficiently,orvieto2023resurrecting,wang2023state}, particularly through the HiPPO theory \cite{gu2020hippo} underlying state space models. HiPPO demonstrates how continuous or discrete sequences can be compressed into high-dimensional vectors via orthogonal polynomial projections. Building on this foundation, we provide a simplified explanatory approach to elucidate the memory capacity of linear dynamical system-based spiking neural networks in the appendix.~\ref{sec:memory}, facilitating reader comprehension.

\begin{figure}[H]
    \centering
    \includegraphics[width=0.75\linewidth]{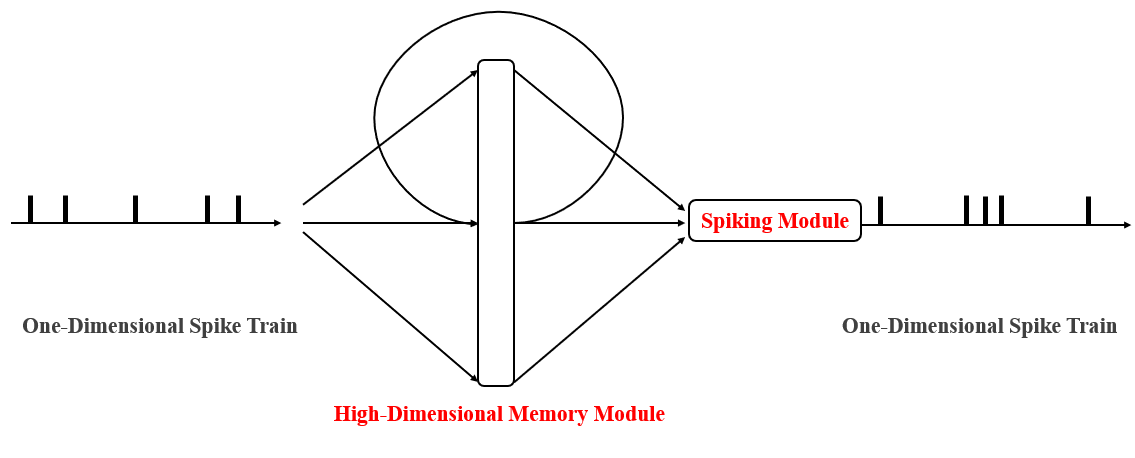}
    \caption{Decoupling Spiking and Memory Modules in Spiking Neural Networks}
    \label{fig:2module}
\end{figure}

Since existing studies have fundamentally explained the nature of sequential memory, we can conclude that: \textbf{for spiking neural networks (SNNs), when disregarding reset mechanisms, they can be viewed as linear dynamical systems with diagonal matrices, where the spiking mechanism merely serves to transmit information between layers without contributing to historical memory storage.} Consequently, current mainstream SNN architectures for sequential tasks can be conceptually divided into two core modules: the \textbf{memory module} and the \textbf{spiking module}.As shown in the Fig.~\ref{fig:2module}.Note that the one-dimensional input here refers to each individual dimension of the high-dimensional input.
Notable implementations include TC-LIF \cite{zhang2024tc}, which utilizes complex-valued state space matrices in multi-compartment structures, and DH-LIF \cite{zheng2024temporal}, employing real-valued state space models with multi-dendrite designs. When decoupled from the spiking mechanism, the memory module becomes functionally equivalent to those used in conventional artificial neural networks, allowing for substitution with various memory systems such as state space models or linear attention mechanisms.This decomposition raises profound questions about the role of biological neural complexity, particularly why the brain employs sophisticated nonlinear dynamics when simpler linear systems appear sufficient for memory tasks. It also prompts reconsideration of how neuromorphic computing should incorporate biological principles. At present, this work remains fundamentally incapable of addressing these questions. 
Therefore, we deliberately shift our research focus away from the memory module, given the demonstrated success of state-space-model-based sequential architectures. This study will instead prioritize analyzing the spiking transmission module in SNN-based sequence tasks, as within this framework, only this component intrinsically involves "spikes." Concentrating on this module is essential for fundamentally understanding the operational nature of spikes in practical applications.

\subsection{Reset Mechanism: A Specialized Discretization Perspective}

Here we focus on the spiking module.The reset mechanism in LIF model has long been a fundamental component of spiking neural networks. While essential for generating sparse spiking patterns through combined use with refractory period functions, reset operations inevitably lead to information loss. Recent studies \cite{fang2023parallel,stan2024learning} have demonstrated successful long-sequence modeling without reset mechanisms, though resulting in denser spike trains. In response, alternative approaches \cite{shen2024spikingssms,zhong2024spike} reintroduce reset mechanisms to maintain sparsity, despite introducing parallelization challenges during training. PMSN~\cite{chen2024pmsn} employs an integrate-and-fire model with soft reset mechanism for spike generation, achieving parallel computation while maintaining sparsity and demonstrating superior performance. However, it lacks theoretical interpretation of the spike sequence mapping,specifically, how such transformation of input sequences should be understood and compared with traditional RNNs or SSM-based ANN sequence models. Crucially, these approaches fundamentally assume that spiking neural networks must incorporate reset and refractory periods following the LIF's real-time updating paradigm. Recent work~\cite{bal2024p} circumvents LIF dynamics entirely through probabilistic spike sampling. Ultimately, if the research objective is practical SNN applications rather than pursuing biologically-plausible intelligence, the biological fidelity of these reset and refractory implementations warrants reconsideration, given they already represent extreme simplifications of neuronal dynamics.

\begin{figure}[H]
    \centering
    \includegraphics[width=0.75\linewidth]{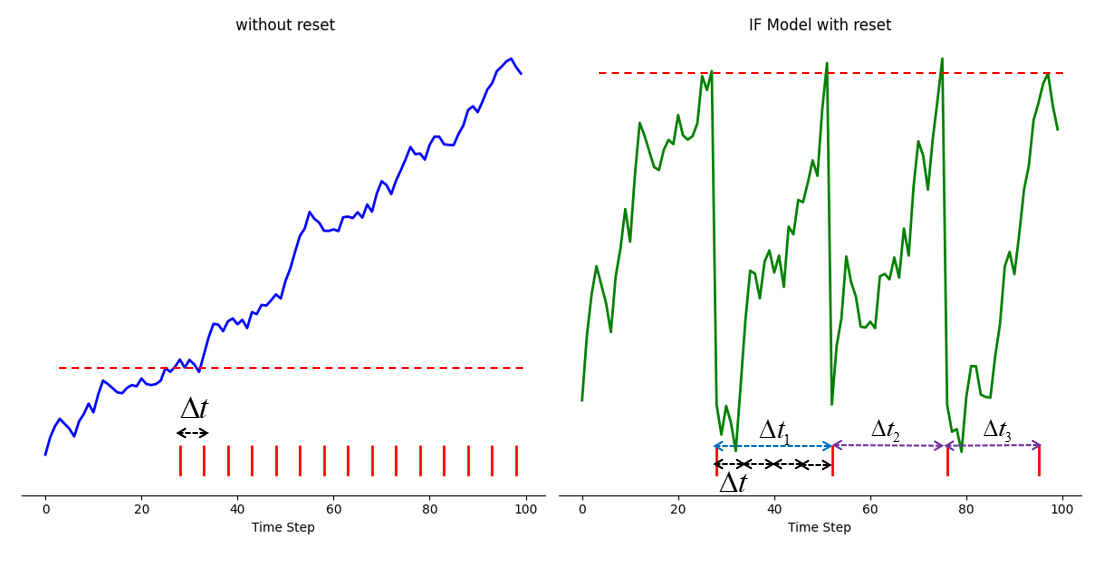}
     \caption{Special Discretization Perspective of Reset Mechanisms}
     \label{fig:resetornot}
\end{figure}

First, let us recall the traditional reset mechanism of spiking neurons, as illustrated in Fig. \ref{fig:resetornot}. For simplicity, the IF model is used here for explanation. The left panel shows the IF model with the reset mechanism removed, while the right panel shows the IF model retaining the hard reset mechanism, with red indicating the spike train. It can be observed that if the reset mechanism is removed, it corresponds to a standard discretized recurrent neural network with binary activation, where each timestep $\Delta t $ remains identical. If the reset mechanism is not removed, it represents a non-standard discretized recurrent neural network. Although inference is performed using $\Delta t$, it is equivalent to employing different discrete timesteps  $\Delta t_1  \neq \Delta t_2 \neq \Delta t_3$.

\begin{figure}[H]
    \centering
    \includegraphics[width=0.75\linewidth]{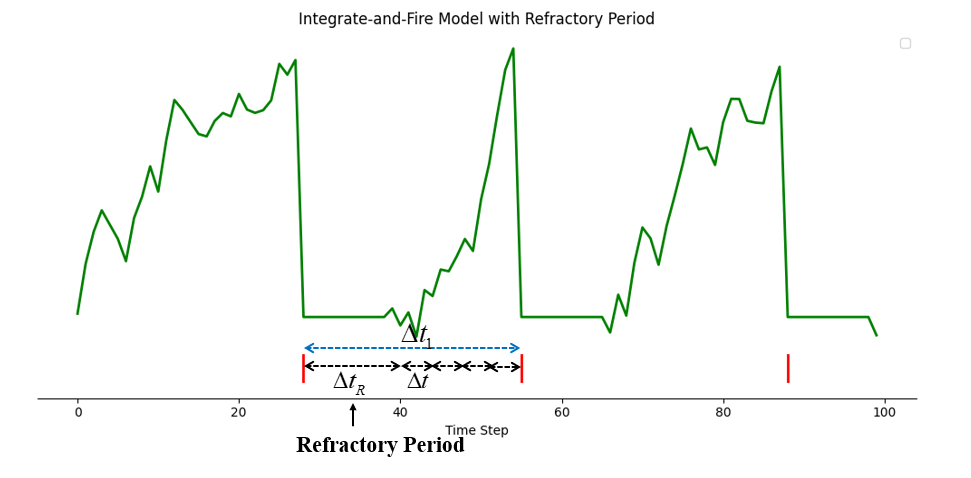}
     \caption{Special Discretization Persperctive of Refractory Period Mechanisms}
     \label{fig:buyinqi}
\end{figure}

Similarly, we further analyze spiking neural networks with refractory periods. As shown in Figure~\ref{fig:buyinqi}, when no spikes are generated, the spiking module of the spiking neural network employs a smaller timestep $\Delta t$ to search for spike timing points. After a spike is generated, the spiking module automatically adopts a larger timestep $\Delta t_R$ (whose duration is unknown). Following this larger timestep, the system resumes using the smaller timestep $\Delta t$ to detect whether the model reaches the threshold. This entire spike-free interval can be considered as a single discrete timestep $\Delta t_1$, as discussed in the previous section.

This leads us to the following theorem:
\begin{theorem}
A spiking neural network can encode the continuous output $f(t)$ of any memory module in the following form:
\begin{equation}
    S[f(t)] =
    \begin{cases}
        1, & \text{if } f(t) \geq \theta \land f(t-\Delta t) \leq \theta \land \dots \land f(t-m(t)\Delta t) \leq \theta \\
        0, & \text{otherwise}
    \end{cases}
\end{equation}
where $S(\cdot)$ denotes the spike encoding function, $m(t) \in \mathbb{Z}^+$ is a time-varying function, and $\Delta t$ represents the discretization timestep.
\end{theorem}

This theorem universally describes the spike encoding scheme for all spiking neural networks with reset mechanisms and refractory periods, where different $m(t)$ values correspond to different reset and refractory configurations. Essentially, the theorem establishes two conditions for spike generation: (1) reaching the threshold and (2) being outside the refractory period. The fundamental obstacle to parallel training lies in the implicit dependence of $m(t)$ on previous outputs. The PMSN model \cite{chen2024pmsn} overcomes this through soft-reset IF neurons that track only spike counts, enabling parallel training. However, LIF models with either hard or soft resists cannot be directly parallelized, prompting alternative solutions like Spikingssm \cite{shen2024spikingssms} and Spikessm \cite{zhong2024spike}.

Traditional spiking neural networks employ encoding schemes where accumulated inputs are transformed into spike trains through LIF or IF models. If we can construct a recurrent neural network with certain discretization mechanisms that simultaneously enables sparse spike generation and parallel training, such networks could serve as alternatives to traditional spiking neural networks. This raises the fundamental question: is it strictly necessary to use LIF or IF models as the pathway for encoding the memory module's output?

The preceding analysis has established that spiking sequences can be interpreted as samples drawn from an underlying distribution. By circumventing traditional IF/LIF encoding schemes and directly propagating the complete distribution shown in Figure \ref{fig:distributionrnn} to subsequent network layers, we gain fundamental insights into the essential nature of inter-spike-sequence mapping,specifically, the simulation of each layer in a real-valued RNN through discrete sampling points that approximate the original continuous distribution. It should be emphasized that this approach differs fundamentally from methods like \cite{bal2024p} that employ population spiking to approximate temporal values, instead implementing the conceptual framework illustrated in Fig.~\ref{fig:distributionrnn}. Such spiking sequences more faithfully preserve the characteristic properties of their RNN counterparts while maintaining closer theoretical alignment with conventional RNNs, thereby offering superior interpretability compared to LIF/IF-based encoding.
Through the model's memory module, we obtain a distribution similar to the output of traditional RNNs. By simply sampling this distribution, the sparse sampling points can characterize the information of the original distribution while simultaneously yielding sparse spike sequences, where the sparsity corresponds to the sampling density. It should be noted that we are not aiming to construct a non-deterministic system, thus eliminating the need for actual probabilistic firing to sample the distribution. We assume either: (1) a dense spike sequence obtained through probabilistic firing based on the output distribution of the memory module, or (2) a dense spike sequence directly generated by threshold detection. Through specialized methods such as introducing a refractory period mechanism,we can then sample this dense spike sequence to obtain the final sparse spike output.

After special discretization, a sparse spike train is obtained, which will deliver spikes at different time steps to neural network layers with identical parameters. That is, whenever a spike with a value of 1 is received, it will have the same impact on the neural network, i.e., the same Post-Synaptic Potential (PSP). Therefore, no matter what kind of refractory function is applied after the spike firing, the spike itself remains an estimate for this time step and exerts an identical influence. For the next layer of the network, the choice of refractory function does not affect the system's perception of the spikes that generate this refractory function. Thus, we can directly set \( m(t) \) as a time-invariant constant, remove the reset mechanism and the accumulation mechanism in Leaky Integrate-and-Fire , and only retain a fixed refractory period to ensure sparse spike firing. This converts the output of the memory module directly into a spike train, resulting in a \textbf{spiking neural network model with a fixed refractory period}, which approximates a continuous-form recurrent neural network (RNN) with irregular time steps. 
The spiking neural network processing spike trains can also be linked to discrete-form RNNs, treating it as a special type of discretized RNN. 

\begin{figure}[H]
    \centering
    \includegraphics[width=0.75\linewidth]{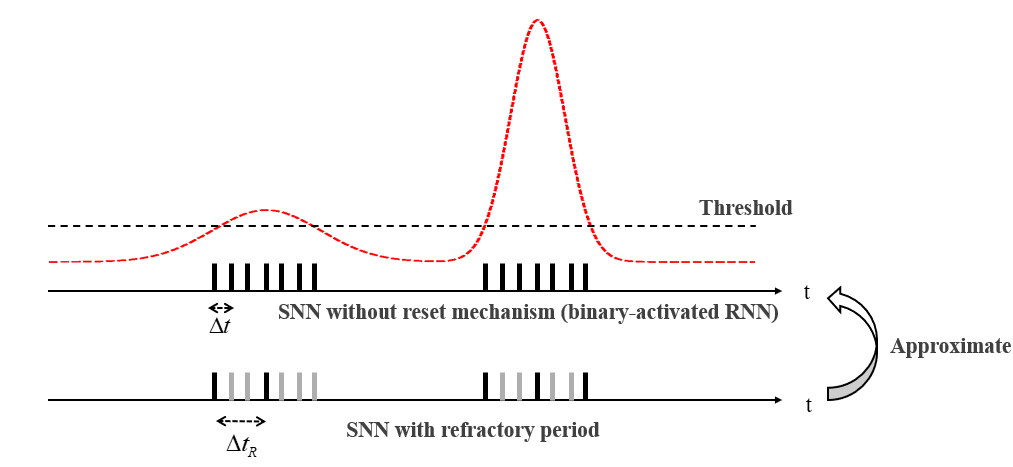}
     \caption{Approximation Perspective of SNNs with Refractory Periods to Conventional Binary RNNs}
     \label{fig:appro}
\end{figure}

Here, each spike serves as an estimate of the output function in the continuous form over neighboring time steps, as shown in Fig. \ref{fig:appro}. The red line represents the output that the memory model needs to pass to the next layer. An SNN without a reset mechanism (i.e., a binary discretized RNN) can be approximated by an SNN with a refractory period. For the latter, whenever the next layer receives a real spike (black), it implies receiving information from multiple spikes, including the subsequent gray ones. Through such sparse spike trains, adding a refractory function can be seen as a special approximation method for binary RNNs, leading to sparser spike sequences. Due to the properties of linear time-invariant systems, if each fired spike carries a different refractory period, the next neural network layer cannot perceive the differences between the received spikes. Therefore, setting the refractory period as a constant also aligns with the characteristics of linear time-invariant systems and enables parallel training of the model.

\begin{figure}[H]
    \centering
    \includegraphics[width=0.75\linewidth]{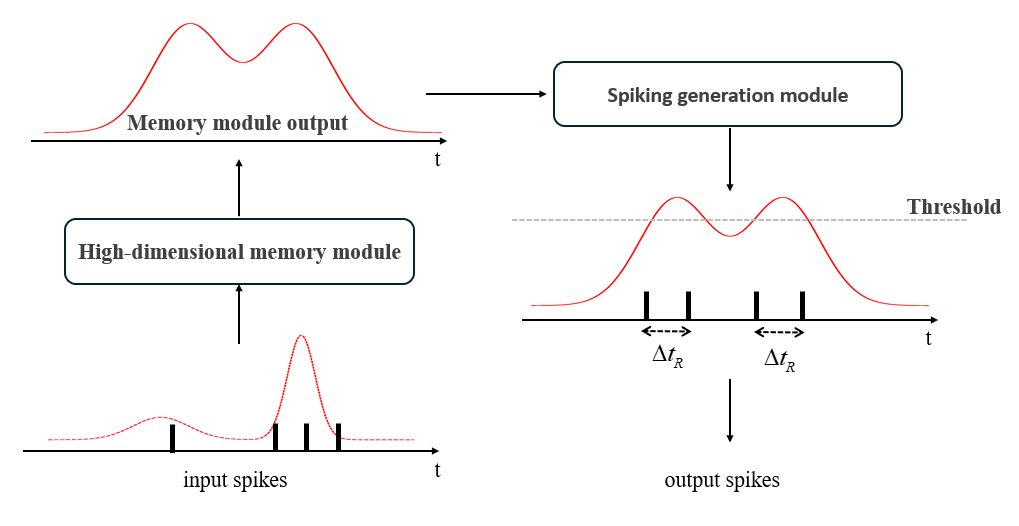}
     \caption{Fixed Refractory Period Spiking Neural Network Memory Output Mechanism}
     \label{fig:Fr}
\end{figure}

Considering continuous-form RNNs, such as Neural ODEs, which receive continuous inputs and pass continuous outputs to the next layer, these outputs can be viewed as a continuous function of some probability distribution. Since the memory module of a spiking neural network is no different from that of a continuous-form recurrent neural network or state-space model, the spike firing mechanism essentially maps such continuous functions into discrete spike trains. It can be argued that the core of SNN sequence mapping is an approximation of continuous functions. By discretizing continuous inputs and Neural ODEs, when the discretization step size is sufficiently small, the model can approximate Neural ODEs. We refer to this discretization approach as a \textbf{regularly discretized recurrent neural network}, while SNNs are termed \textbf{irregularly discretized recurrent neural networks}.  
Here, we summarize the overall characteristics of the model: \textbf{The sequence mapping of spiking neural networks can essentially be viewed as updating real-time memory for sequences and emitting spikes at irregular discrete time steps, approximating the output distribution of traditional continuous-form RNNs in the form of sampled points. The use of reset mechanisms and refractory periods constitutes this special discretization approach, distinguishing it from traditional RNNs with regular discrete time steps.} The overall concept is illustrated in Fig. \ref{fig:Fr}.

With the completion of the core theoretical framework, we now proceed to construct a parallelizable spiking neural network based on state-space models in the following section.

\subsection{Model Architecture}
This section proposes two spiking neural network models based on state-space models. The first is the simplest form of a fixed refractory period-based spiking neural network. While this model's practical performance is suboptimal, its conceptual framework may enhance our understanding of spiking neural networks. Readers not interested in this approach may skip directly to the second subsection. The second section re-examines the essence of sparse spiking and introduces the second model, \textit{spikingPssm}, which is essentially a state-space model combined with a special form of PSN \cite{fang2023parallel}. The model remains equally simple, but it should be noted that such a model may offer better interpretability. If a sufficiently simple model can achieve satisfactory results without requiring other complex mechanisms, it may be more likely to be adopted in practical applications. Following the state-space model approach, each input dimension is processed separately through the memory module to produce outputs of corresponding dimensionality, which are then fed into the spiking module, nonlinear activation modules (MLP, GLU, etc.). Therefore, all one-dimensional inputs in subsequent diagrams refer to individual dimensions of the input signal processed separately.

\subsubsection{Fixed Refractory Period-based Spiking Neural Network (spikingFRssm)}
SpikingFRssm combines a memory module (state-space model) with a spiking module employing a fixed refractory period strategy for parallel training and enhanced interpretability. Like standard state-space models, it processes each input dimension separately through the memory module before passing outputs to nonlinear activation modules (MLP, GLU, etc.). The memory module implementation follows established work \cite{gu2021efficiently,smith2022simplified,orvieto2023resurrecting} and thus requires no further elaboration here.

The key differentiators from SpikeSSM \cite{zhong2024spike}, spikingSSM \cite{shen2024spikingssms}, PMSN \cite{chen2024pmsn}, and TC-LIF \cite{zhang2024tc} lie in the spiking mechanism. While the memory module remains consistent with these models (all similar to S4 architecture), our approach introduces the followed components:

\textbf{Spike Activation Function}

The spiking module's core function is generating sparse spike trains while preventing continuous firing upon reaching threshold potentials. Based on our theoretical analysis, we implement a fixed refractory period mechanism. First of all we need to explain the spike activation function. Unlike approaches that feed Memory module's outputs as currents into LIF/IF models, we directly treat memory module's outputs as membrane potentials, similar to \cite{stan2024learning}. For clarity, we denote inputs from memory module as $x$, with the spiking activation function defined by Equation (\ref{spike}):

\begin{equation}
s(t) = H[x(t)-\theta]=
\begin{cases} 
1, & \text{if } x(t) > \theta \\
0, & \text{else}
\label{spike}
\end{cases}
\end{equation}
where $H(\bullet)$ represents the Heaviside step function.

\textbf{Linear Refractory Function}

After generating spikes , the model enters a refractory period implemented through a linear function applied directly to input $x$ (without reset) to preserve information and enable parallel training (detailed later):

\begin{align}
    \eta(t+t') &= -m n + m t', \quad t' = 0, 1, 2, \dots, n\\
    m &= \max_{0 \leq \tau \leq t} x(\tau)
\end{align}
Here, $n$ denotes the fixed refractory duration, while $m$ can be any sufficiently large value ensuring inputs remain subthreshold during refractory periods. This linearly increasing function controls spiking frequency without additional computational overhead, producing sparse spike trains. During inference, when membrane potential reaches threshold at time $t$, the refractory function modifies subsequent inputs without resetting them,unlike PMSN and TC-LIF, our model retains complete input information, improving interpretability.

\textbf{Surrogate Gradient Function}

\begin{figure}
    \centering
    \includegraphics[width=0.5\linewidth]{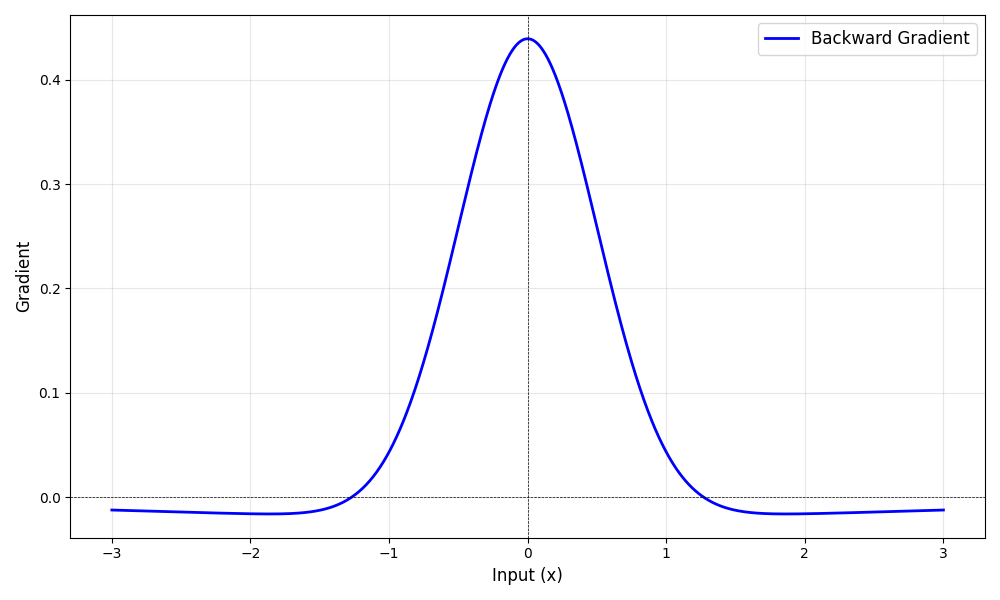}
     \caption{Surrogate gradient function used in our study}
     \label{fig:grad}
\end{figure}

To address the non-differentiability of spike generation, we employ the surrogate gradient function \cite{yin2021accurate} shown in Fig. \ref{fig:grad}:

\begin{equation}
    G(x) =  \gamma \cdot \left[ (1 + h) \cdot \phi(x; 0, \sigma) - h \cdot \phi(x; l, \sigma') - h \cdot \phi(x; -l, \sigma') \right]
\end{equation}
where $\phi(x; \mu, \sigma) = \dfrac{1}{\sigma \sqrt{2\pi}} e^{-\dfrac{(x - \mu)^2}{2\sigma^2}}$ is the Gaussian function, with parameters $h = 0.15$, $\sigma = l$, $\sigma' = 6l$, and $l=\gamma=0.5$. During backpropagation, $G(\bullet)$ substitutes for the non-differentiable $H(\bullet)$ (e.g., the derivative of $H(x-\theta)$ becomes $G(x-\theta)$). The other alternative surrogate gradients may also be used.

\textbf{Parallel Training Inference Method}

During the training phase, the spiking module is processed in parallel training manner following the parallel-trained memory module. For inference, the sequential computation is performed using the linearly increasing refractory function described earlier, while during training, a novel computation approach is adopted.
The essential purpose of employing the linear refractory function is that when the membrane potential reaches the threshold and emits a spike at a certain moment, it inhibits subsequent spike emission for a period of time. In the model of this chapter, this duration is fixed, thus parallel training through fixed-length convolution can also be considered. The specific approach is as follows: when determining whether to emit a spike at a given time step, check whether spikes were emitted in the preceding r-1 time steps. The concrete method is: first pass the output of the memory module through a spiking activation function (step function):

\begin{equation}
    y' = H(y-\theta)
\end{equation}
where $\theta$ is the spiking threshold and $y(t)$ the input at time $t$. This produces unmodified spike trains without refractory effects. We then apply a fixed convolutional kernel:

\begin{equation}
\begin{aligned}
\text{kernel} &= \underbrace{[-2.0, -2.0, \ldots, -2.0,}_{\text{r} - 1 }  1.0] \\
\end{aligned}
\end{equation}

This 1D kernel operates on each dimension $y'_i$ of $n$-dimensional input $y'$:

\begin{equation}
y''_i(t) =(kernel * y_i')(t)= \sum_{n=0}^{r-1} y'_i(t-n) \cdot kernel(r-n)
\end{equation}

Defining $q$ as the count of elements in time window $[t-r, t)$ satisfying $y''_i(t') > \theta$:

\begin{equation}
    q = \sum_{t' \in [t-r, t)} \mathbb{I}(y''(t') > \theta)
\end{equation}

When $q > 0$, the convolution result satisfies:

\begin{equation}
    y''(t) = -2q+1 < 0
\end{equation}

The final spiking module output becomes:

\begin{equation}
    output_i = y' \odot H(y'')
\end{equation}

\begin{figure}
    \centering
    \includegraphics[width=0.75\linewidth]{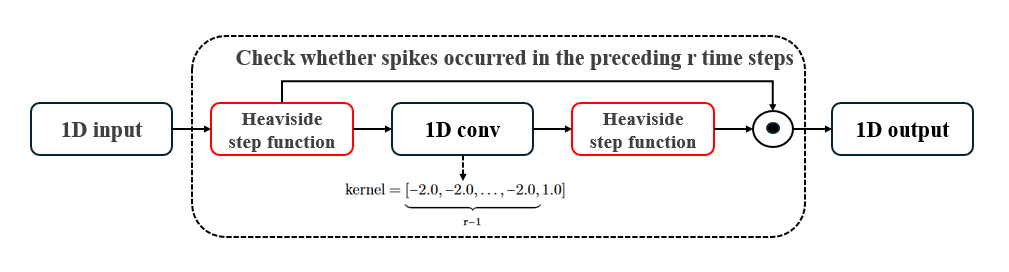}
    \caption{Parallel training mechanism for SpikingFRssm's spiking module}
    \label{fig:sfr}
\end{figure}

Here, $y'$ represents direct memory module outputs, while $H(y'')$ acts as a temporal mask suppressing dense spikes to produce sparse trains (Fig. \ref{fig:sfr}).

By adopting this computational approach to replace the sequential inference method with linear refractory functions, identical spike output results can be obtained. Therefore, during inference, the linear refractory scheme can be employed for sequential processing with linear complexity, while during training, parallel processing of all timesteps can be implemented to significantly enhance training efficiency.

\subsubsection{Regularly Discretized Spiking Neural Network (spikingPssm,P:Parallel)}
Sparse spiking implies maximizing silent intervals,a key challenge in current spiking models. But must outputs be strictly 0 for energy efficiency? If all timesteps in an interval share identical outputs, we could: (a) merge subsequent inputs at the first timestep during deployment while keeping later steps silent, or (b) store the first step's output to skip computations. Both approaches achieve energy savings comparable to traditional sparse spiking. We first derive:

\begin{theorem}
For a state-space model $\frac{dy}{dt}=Ay+Bx$, when inputs $x(t),t\in[0,r]$ remain constant over interval $r$, $y(r)$ equals the output at $t=0$ with input:
\begin{equation}
    x(0)\frac{\int_0^r e^{A (r-\tau)} B \, d\tau}{\int_0^1 e^{A (r-\tau)} B \, d\tau}
\end{equation}
\label{x0=xt}
\end{theorem}
Proof: The general solution is:
\begin{equation}
y(t) =  e^{A t} y(0) + \int_0^t e^{A (t-\tau)} B x(\tau) \, d\tau
\end{equation}
For constant $x(t)\in[0,r]$:
\begin{equation}
    \int_0^r e^{A (r-\tau)} B x(\tau) \, d\tau=x(0)\int_0^r e^{A (r-\tau)} B \, d\tau
\end{equation}
For $x'(t)$ constant in $[0,1]$ and zero in $(1,r]$:
\begin{equation}
     \int_0^r e^{A (r-\tau)} B x'(\tau) \, d\tau=x'(0)\int_0^1 e^{A (r-\tau)} B \, d\tau
\end{equation}
Equivalence holds when $x'(0)=x(0)\frac{\int_0^r e^{A (r-\tau)} B \, d\tau}{\int_0^1 e^{A (r-\tau)} B \, d\tau}$.

Therefore, it is entirely feasible to construct a sparse spiking neural state-space model using regular discrete time steps. Only a single computation of the GLU module's output is required, which when multiplied by the weight in Theorem \ref{x0=xt} effectively integrates values across the entire interval. This also corresponds to the theoretical perspective of viewing spiking neural networks as discrete approximations of continuous binary recurrent neural networks.
Alternatively, We may  adopt another implementation approach: only the spiking pattern at the initial timestep needs to be computed. For subsequent intervals with identical spiking patterns, no further computation is required,simply preserving the initial spiking pattern suffices. The resulting storage complexity is equivalent to maintaining membrane potentials in LIF models.

SpikingPssm retains the standard memory module from S4 models. For spiking generation, inspired by PSN \cite{fang2023parallel}, we implements a temporal convolutional approach with output sharing across timesteps. For each dimension $x(t)$ from the memory module, given refractory length $r$ (better termed temporal kernel size), the membrane potential is:

\begin{equation}
\begin{aligned}
     u(t)&=\sum _{i=a}^bw(i-a)x(i)\\
     a&=r\left\lfloor \frac{t}{r} \right\rfloor\\
     b&=r\left\lceil \frac{t}{r} \right\rceil
\end{aligned}
\end{equation}
where $\left\lfloor \bullet \right\rfloor$ and $\left\lceil \bullet \right\rceil$ denote floor/ceiling operations, and $w=[w_1,w_2\dots w_r]$ are learnable parameters. Membrane potentials are shared across $[a,b]$. Spike generation follows:
\begin{equation}
    output(t)=H(u(t)-\theta)
\end{equation}

\begin{figure}[H]
    \centering
    \includegraphics[width=0.75\linewidth]{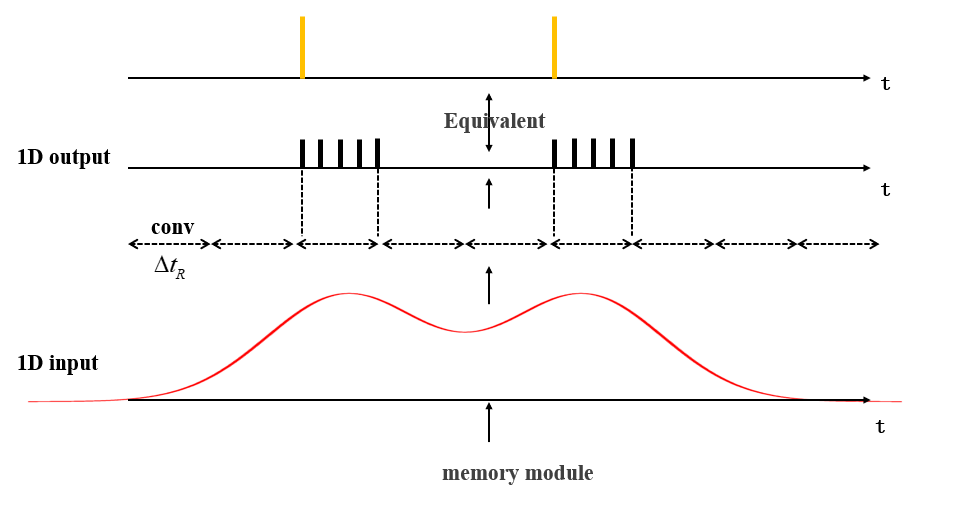}
     \caption{SpikingPssm spiking module architecture}
     \label{fig:Pssm}
\end{figure}

Figure \ref{fig:Pssm} illustrates the spiking module. Compared to PSN, SpikingPssm incorporates state-space memory while employing a specialized kind of sliding PSN \cite{fang2023parallel} for spiking module.The model implementation is extremely simple, see the Appendix.~\ref{sec:pssm} for details.

\section{Experiments}
We directly evaluate the model's sequential modeling capability on the Sequential CIFAR-10 dataset. Given the model's relatively simple architecture and our primary focus on analyzing the fundamental nature of reset or refractory mechanisms, we primarily aim to demonstrate that even such a basic structure can achieve competitive results, rather than pursuing state-of-the-art performance. Consequently, we specifically highlight the experimental results obtained on this single dataset.

\subsection{Dataset Description}
Sequential CIFAR10 adapts the CIFAR-10 dataset by flattening 32$\times$32 grayscale images into 1024-length pixel sequences, transforming image classification into sequence modeling. Retaining the original 50K training and 10K test samples, this variant specifically tests models' ability to process ultra-long sequences, requiring classification after progressively receiving all pixels.

\subsection{Experimental Configuration}
Our state-space model implementation builds upon s4d \cite{gu2022parameterization,gu2021efficiently}. Table \ref{tab:hyperparams_compact} details the parameter settings for comparative experiments. Both SpikingFRssm and SpikingPssm share identical configurations, with SpikingPssm's convolutional kernel length equating to its refractory period. Notably, spiking thresholds differ: 0.5 for SpikingFRssm versus 0 for SpikingPssm.

\begin{table}[h]
\centering
\caption{Model Hyperparameters}
\begin{tabular}{l c @{\qquad} l c}
\toprule
\textbf{Parameter} & \textbf{Value} & \textbf{Parameter} & \textbf{Value} \\
\midrule
Learning rate & 0.001 & Reset timestep & 5 \\
Weight decay & 0.01 & Layers & 6 \\
Epochs & 200 & Dropout & 0.1 \\
Kernel dim & 64 & Model dim & 512 \\
\bottomrule
\end{tabular}
\label{tab:hyperparams_compact}
\end{table}

\subsection{Comparative Analysis}
Results demonstrate SpikingPssm's significant superiority over SpikingFRssm. This likely stems from: (1) SpikingPssm's learnable convolutional kernel capturing local input features, and (2) full utilization of temporal information during training. While SpikingFRssm conceptually estimates dense spikes with discrete sequences, its fixed convolutional kernel masks partial information,similar to reset mechanisms,impeding complete information flow. Making SpikingFRssm's kernel learnable achieves comparable performance to SpikingPssm, but sacrifices sparse spiking by essentially adding a local feature extraction module after binaryS4\cite{stan2024learning}.

Through comparison with other SSM-based spiking neural network models, it can be observed that SpikingPssm outperforms P-SpikeSSM models that employ random sampling for spike generation and PMSN models that use IF as the spike generation function. SpikingPssm also outperforms GSU models that completely eliminate reset mechanisms and adopt pure binary activation. The reason is that GSU performs quantization processing in the GLU module, resulting in lower power consumption. Here, we do not conduct in-depth research on additional incremental lightweight techniques such as quantization, but focus solely on the simplest form of spike generation. SpikingPssm employs 
 a special kind of PSN\cite{fang2023parallel} for spike generation, which can also be viewed as local information enhancement, but it underperforms compared to SpikingSSM and SpikeSSM. Although the model in this study is not SOTA, our method does not introduce additional complex computational approaches. The achieved results can, to some extent, demonstrate the effectiveness of this spike generation mechanism. Moreover, this model exhibits stronger interpretability, closely aligning with the learning paradigm of binary-activated RNNs, which contributes to a deeper understanding of the essence of spiking neural networks. Since all these SSM-based models perform better than non-SSM architectures such as PSN, LSNN\cite{bellec2018long}, ALIF\cite{yin2021accurate}, etc., and also outperform TC-LIF that employs 2D-SSM, this demonstrates the superiority of our model compared to these other types of models. It also illustrates the sequence modeling capability of multi-compartment, multi-dendrite SSM models, further confirming that spiking is independent of memory mechanisms, and shows that the output of the memory module can transmit most information through the spike generation method proposed in this study.
\begin{table}[htbp]
\centering
\caption{Comparative Results}
\label{tab:snn_comparison}
\begin{tabular}{lllS[table-format=2.2]}
\toprule
  \textbf{Method} & \textbf{Training} & \textbf{Params(K)} & \textbf{Accuracy (\%)} \\
\midrule
 SPSN\cite{fang2023parallel} & Parallel & 184 & 70.2 \\
 PMSN\cite{chen2024pmsn} & Parallel & 215 & 82.1 \\
GSU\cite{stan2024learning}& Parallel & N/A & 85.0\\
P-spikessm\cite{bal2024p}& Parallel& N/A & 82.4\\ 
Spikessm\cite{zhong2024spike}& Parallel& N/A& 87.2 \\
Spikingssm\cite{shen2024spikingssms}& Parallel& N/A & 86.8 \\
\midrule
spikingFRssm & Parallel & 359 & 80.0 \\
spikingPssm & Parallel & 359 & 85.5 \\
\midrule
\label{compare}
\end{tabular}
\end{table}

\subsection{Energy Efficiency Analysis}
This section will analyze and compare the energy consumption of SpikingPssm. Here, we do not focus on the floating-point operations of the memory module. Under the current state-space-model-based framework, since the floating-point multiplications caused by the recurrent structure of the memory module are unrelated to spikes themselves, this paper does not deeply investigate the power consumption brought by the memory module. Similar to SpikingSSM\cite{shen2024spikingssms} and SpikeSSM\cite{zhong2024spike}, we focus here on the sparse spike representation after the memory module. Sparse spike generation can reduce the number of computations in the channel-mixer layer's GLU or MLP after the memory module, meaning computations are only performed when spikes occur. This is also the energy advantage compared to traditional ANNs.

Note that in SpikingPssm, our definition of spike sparsity differs,it does not refer to the actual spike firing ratio. What we call "fixed refractory period" here is essentially the length of the temporal convolution kernel. It is termed "fixed refractory period" because computations need only be performed once at the first timestep, with no further computations required for the entire duration of this length. When the refractory period is longer, the spike firing frequency is more likely to decrease. For example, with a refractory period length of 5, the actual computation frequency can be considered as the true spike firing frequency divided by 5, since the spike firing patterns remain identical across these five timesteps; with a refractory period length of 3, it is equivalent to the true spike firing frequency divided by 3.

In LIF models, the membrane potential at each timestep is multiplied by a constant corresponding to membrane potential coefficient $\tau$; whereas under SpikingPssm's spike generation mode, the output of the memory module at each timestep is multiplied by a parameter determined by the temporal convolution kernel. Thus, the computational complexity of both methods is identical. Moreover, the computational complexity remains the same for different refractory period lengths,each timestep involves multiplication by one parameter.

\begin{figure}[htbp]
    \centering
    \begin{minipage}{0.48\textwidth}
        \centering
        \includegraphics[width=\linewidth]{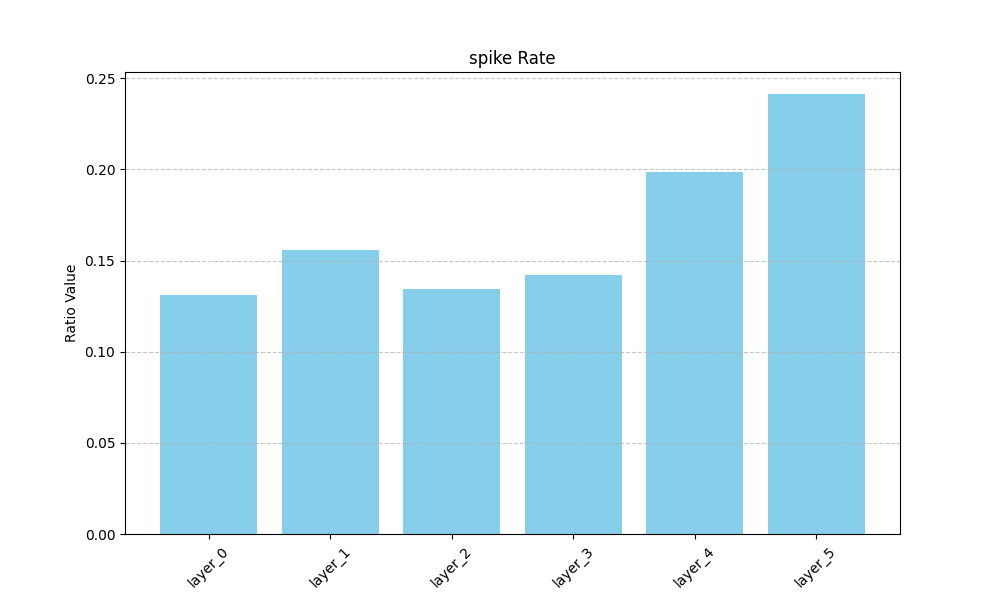}
        \caption{Spiking frequency (Refractory=5)}
        \label{fig:3a}
    \end{minipage}
    \hfill
    \begin{minipage}{0.48\textwidth}
        \centering
        \includegraphics[width=\linewidth]{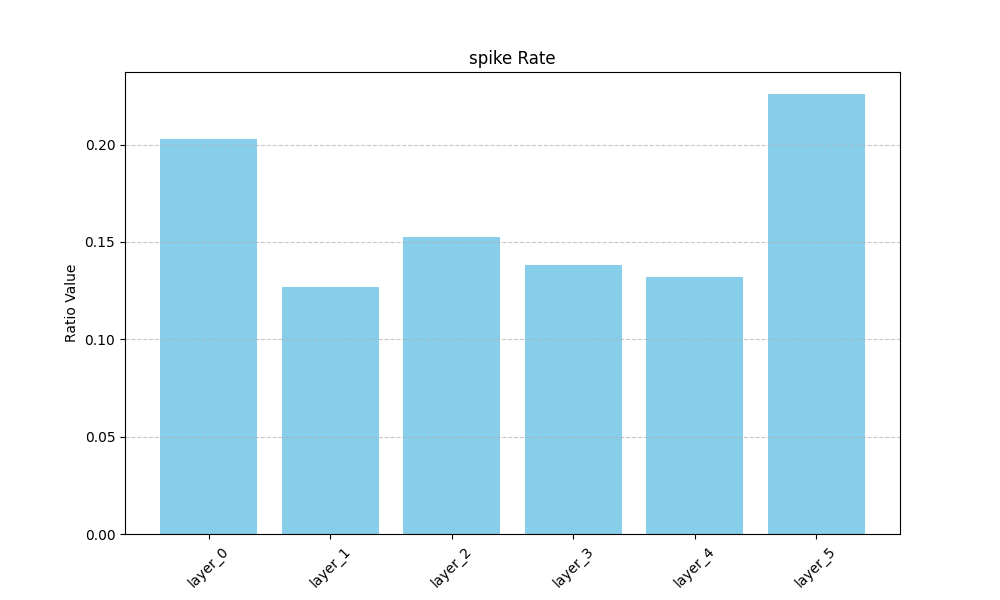}
        \caption{Spiking frequency (Refractory=3)}
        \label{fig:3b}
    \end{minipage}
    
\end{figure}

In the SCIFAR task, the actual spike firing frequency for each layer is as shown in the Fig.~\ref{fig:3a} and the Fig.~\ref{fig:3b}. When the refractory period length is 5, the actual spike firing frequency is approximately 0.15, making the computed spike frequency about 0.03; when the refractory period length is 3, the actual spike firing frequency is also approximately 0.15, making the computed spike frequency about 0.05. Future research could further explore how to reduce the computational cost of local information extraction and memory module operations.

\begin{figure}[H]
    \centering
    \includegraphics[width=0.5\linewidth]{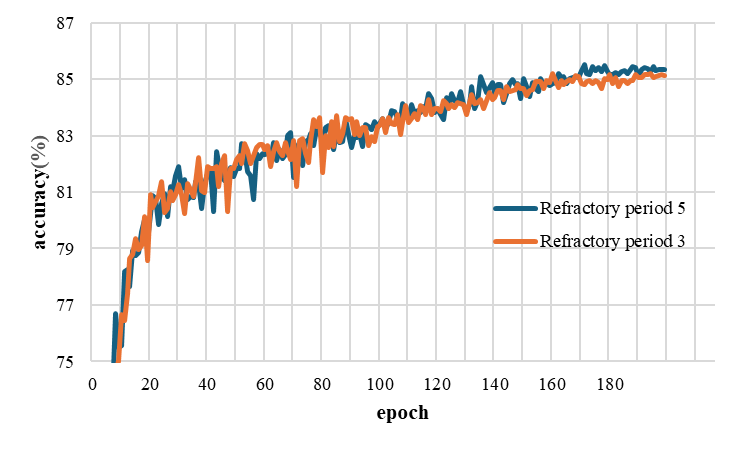}
    \caption{Impact of refractory period settings}
    \label{fig:peroid}
\end{figure}

Here we analyze the impact of refractory period length on SpikingPssm's experimental results. The accuracy change curves for both cases are shown in Figure \ref{fig:peroid}. The model with refractory period 5 slightly outperforms the one with refractory period 3.
We cannot definitively determine the actual effect of refractory period length on task performance. Longer refractory periods provide larger local receptive fields but reduce spike sparsity, potentially weakening the model's representational capacity. Conversely, shorter refractory periods yield smaller receptive fields while maintaining higher spike firing rates to preserve representational capability.

\section{Conclusion}
In this study, we address the following questions:

1. Understanding the reset and refractory mechanisms from the perspective of binary-activated recurrent neural networks. The reset and refractory mechanisms can be interpreted as a specialized form of discretization.

2. Whether to use the reset mechanism or refractory mechanism and how to understand their usage? If the output pattern of the memory module is viewed as a distribution (even though the values of this "distribution" may exceed 1), and the spike train is regarded as sampling points from such a distribution, then in practical applications, it is entirely possible to retain the membrane potential of the model without resetting it but instead using a refractory period. This approach is equivalent to a further sampling method for densely firing spike trains, thereby obtaining sparse spikes. Such a spiking pattern allows spiking neural networks to be fully understood through discrete binary-activated recurrent neural networks, offering stronger interpretability. The essence of information transmission between layers in spiking neural networks also becomes easier to comprehend,it is equivalent to traditional recurrent neural networks or state-space models, where spike trains serve as sampling points to approximate the continuous-form information transmission between layers in recurrent neural networks.

3. We propose spikingPssm, a parallel-trainable sparse spiking neural network state-space model. Its essence can be seen as attaching a specialized PSN \cite{fang2023parallel} for spike generation after an SSM. Although this approach lacks innovation and merely stitches together two existing modules, its simplicity still achieves experimental results close to the best. This demonstrates the effectiveness of adopting such a spike generation module, suggesting that perhaps we do not need to use LIF or IF, which may be somewhat difficult to understand in practical applications.

The model still has limitations. We did not genuinely investigate so-called low power consumption. The complex transformations of state vectors in the memory module still involve intricate floating-point operations. The core significance of this study lies in understanding the essence of spikes in sequential tasks. We hope this research can offer readers a little inspiration.
\bibliography{bibliography.bib}
\bibliographystyle{unsrt}
\clearpage  
\section*{\centering Appendix}  
\setcounter{section}{0}
\section{Declarative Network Perspective of Two-Phase TTFS-based Spiking Neural Networks}
\label{sec:ttfs}
This subsection is not directly related to the main content of this article but serves as an example of the quantized activation value perspective in spiking neural networks. Additionally, my previous paper\cite{zhang2024research} had some shortcomings in terms of rigor and detail, so I revisit and present its core content in Appendices~\ref{sec:ttfs} and~\ref{sec:memory} through a more detailed analysis.
Appendix~\ref{sec:ttfs} shows that if a spiking neural network employs separate timesteps for each layer, such as in the two-phase Time-to-First-Spike (TTFS) scheme, it can, in essence, be functionally equivalent to many quantized declarative artificial neural networks.
Appendix~\ref{sec:memory} offers a straightforward, distribution-based perspective to understand the fundamental nature of information encoding in spike trains. 

Two-stage TTFS-based spiking neural networks refer to networks where each layer operates in a separate time phase, ensuring that each neuron in each layer fires at most one spike. However, this requires longer inference time steps because each layer needs its own dedicated phase for computation. Previously discussed rate-coded spiking neural networks can be easily understood through the lens of frequency as quantized activations in traditional neural networks. Here, we further explain the declarative network nature of temporally coded spiking neural networks—specifically, two-phase TTFS-based spiking neural networks,which have achieved strong performance, helping readers better understand the relationship between spiking neural networks and quantized neural networks.  
For readers interested in the specifics of two-stage TTFS-based spiking neural networks, please refer to literature such as \cite{stanojevic2024high,wei2023temporal,yang2023lc}. These models typically consist of two phases: the first phase receives spike inputs, often using a linear increase in membrane potential with a slope determined by parameter W, followed by the second phase, the spike generation phase, where the neuron fires its only spike once the linearly increasing membrane potential reaches the threshold. We will not delve into further details here. Instead, we aim to highlight that this second phase can be viewed as a step-by-step search for the solution to an equation, thus making it interpretable as a declarative network\cite{gould2021deep} with quantized activations.

Declarative networks \cite{gould2021deep} differ from conventional feedforward networks by defining layer outputs as solutions to optimization problems. Given a parameterized objective \( f : \mathbb{R}^n \times \mathbb{R}^m \to \mathbb{R} \) and constraint set \( C \subseteq \mathbb{R}^m \), the output \( y \) solves:
\[
y \in \operatorname*{arg\,min}_{u \in C} f(x, u), \tag{7}
\]
where \( x \in \mathbb{R}^n \) represents parameters and \( u \in \mathbb{R}^m \) optimization variables. This framework encompasses models like Deep Equilibrium Networks (DEQ) \cite{bai2019deep} and Neural Memory ODE \cite{yi2023nmode}.

Spiking neural networks that emphasize temporal information and employ temporal encoding can also be viewed as a form of declarative network\cite{nilsson2023integration}. These models focus on spike timing, where the firing times of neurons in each layer encode the layer's information. The key question then becomes: when should each neuron in a layer fire? Thus, temporally encoded spiking neural networks can be interpreted as solving an equation to compute the spike timing of neurons in each layer as its output. Specifically, they solve the equation $f(x,y,\theta)=0$, where $y$ is the variable to be solved (i.e., the spike timing), $x$ represents the input spike timing, and $\theta$ denotes the learnable parameters.

Taking the simplest feedforward network $y=ReLU(wx)$ as an example, this is equivalent to solving the equation $y-ReLU(wx)=0$, where the solution is obtained through numerical computation, advancing one time step at a time.

An important constraint is that the equation being solved must be non-coupled. That is, for a declarative network $y=f(W_1y+W_2x+b)$,for the desired output vector $y$, each dimension must be independent, meaning the parameters $W_1$ preceding $y$ must form a diagonal matrix. Only under this condition can a two-phase TTFS-based spiking neural network solve the equation within a fixed time window. If it is not a diagonal matrix, then models like DEQ would be required, which involve iterative computation to solve the equation for each layer's output.

\begin{figure}[H]
    \centering
    \includegraphics[width=0.75\linewidth]{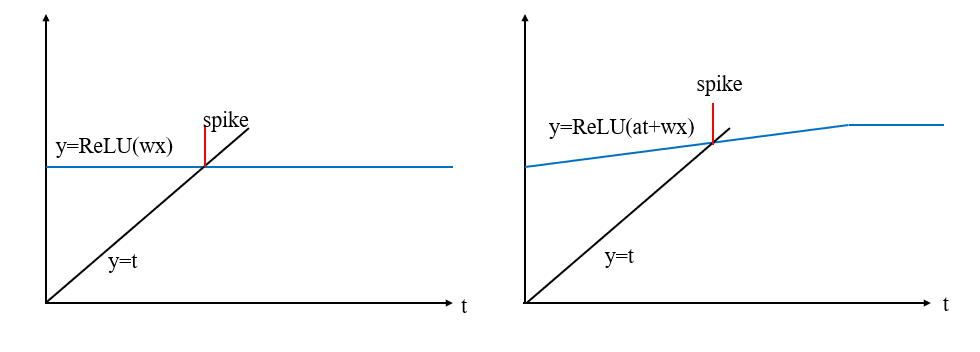}
    \caption{Generalized view of TTFS-SNN firing phases: (left) solving \( y=ReLU(wx) \), (right) solving \( y=ReLU(ay+wx) \)}
    \label{fig:ttfs}
\end{figure}

We can further generalize this perspective. The ReLU(wx) can be replaced with other activation functions or specific formulations, i.e., $f(wx)$, or even state space model expressions like $Au+wx$, where the $Au$ operation can be implemented similarly to $wx$ using a first phase with slope $A$. From this viewpoint, any non-coupled declarative network with quantized activations can be simulated by two-phase TTFS-based spiking neural networks. For instance, solving the output of equations like $y=ReLU(ay+wx)$, etc., as shown in Figure \ref{fig:ttfs}. 

In summary, the second phase of two-phase TTFS-based spiking neural networks, which solves for spike timing, can be viewed as a step-by-step solution process for non-coupled equations. This perspective potentially enables the extension to more types of neural network models.

Thus, both temporally-coded (TTFS-based) and frequency-coded SNNs fundamentally connect to quantized activation ANNs,the former through declarative equation solving, the latter through rate-based quantization.
\section{Memory Mechanisms in Memory Modules}
\label{sec:memory}
The memory mechanisms of linear recurrent neural networks have been extensively studied. For in-depth understanding, readers may refer to \cite{gu2020hippo,orvieto2023universality,wang2023state,smith2022simplified,orvieto2023resurrecting} or mathematical concepts like Fourier transforms and wavelet analysis. These works thoroughly explain the memory essence of current mainstream models. Practical brain-inspired mechanisms for long-sequence modeling, such as multi-compartment models \cite{chen2024pmsn}, also relate to these foundations. This section provides a simplified explanation of sequence memory from the perspective of spiking as sampling points, following Hippo theory \cite{gu2020hippo}.

A probability distribution is uniquely determined by its complete set of moments (first-order to infinite-order). Thus, we can represent distribution $P$ using an infinite-dimensional vector $p=(E(x),E(x^2),E(x^3)\dots,E(x^n)\dots)$. However, practical implementations require finite-dimensional representations. Similar to Hippo's approach, we project the infinite vector onto finite dimensions, e.g., $\hat{p}=(E(x),E(x^2),E(x^3)\dots,E(x^n))$. The challenge lies in efficiently computing these moments for each spike at time $t_k$, as direct computation would involve: (1) excessive floating-point operations, and (2) high computational complexity from cumulative multiplications.

The exponential function in reset-free LIF models offers an efficient alternative. Consider the Taylor expansion:
\begin{equation}
    e^x=1+x+x^2/2+...x^n/n!+\dots
\end{equation}
Projected to n-dimensional polynomials:
\begin{equation}
    proj(e^x)=1+x+x^2/2+...x^n/n!
\end{equation}
We establish the relationship through:
\begin{equation}
\begin{pmatrix}
1&\theta_0&\frac{\theta_0^2}{2}&\frac{\theta_0^3}{3!}&\cdots&&\frac{\theta_0^n}{n!}\\
1&\theta_1&\frac{\theta_1^2}{2}&\frac{\theta_1^3}{3!}&\cdots&&\frac{\theta_1^n}{n!}\\
1&\theta_2&\frac{\theta_2^2}{2}&\frac{\theta_2^3}{3!}&\cdots&&\frac{\theta_2^n}{n!}\\
\vdots&\vdots&\vdots&\vdots&\cdots&&\vdots\\
1&\theta_n&\frac{\theta_n^2}{2}&\frac{\theta_n^3}{3!}&\cdots&&\frac{\theta_n^n}{n!}\\
\end{pmatrix}
\cdot 
\begin{pmatrix}
1\\
t_1\\
t_1^2\\
\vdots\\
t_1^n\\
\end{pmatrix}\approx
\begin{pmatrix}
e^{\theta_0t_1}\\
e^{\theta_1t_1}\\
e^{\theta_2t_1}\\
\vdots\\
e^{\theta_nt_1}\\
\end{pmatrix}
\end{equation}
This shows $f(t_i)=(t_i,t_i^2,...t_i^n)^T$ can be linearly represented by $E(t_i)=(e^{\theta_1t_i},e^{\theta_2t_i}\cdots e^{\theta_nt_i})^T$.

To clarify the exponential function's role, consider a simplified LIF model without reset or refractory periods. For one-dimensional input, the membrane potential evolves as:
\begin{equation}
    V(t) = V(0) e^{-t / \tau_m}
\end{equation}
where $\tau_m$ is the membrane time constant. When receiving a spike $s(t_k)$ at time $t_k$:
\begin{equation}
    V(t_k) = V(t) e^{-(t_k-t) / \tau_m}+ws(t_k)
\end{equation}

Extending to $n$ LIF neurons with different time constants yields the solution to a high-dimensional linear differential equation:
\begin{equation}
    \begin{pmatrix}
        V_1(t_k) \\
        V_2(t_k) \\
        \vdots \\
        V_n(t_k)
    \end{pmatrix}
    =
    W S_{t_1}+
    \begin{pmatrix}
        V_1(t) \\
        V_2(t) \\
        \vdots \\
        V_n(t)
    \end{pmatrix}
    \odot
    \begin{pmatrix}
        e^{-(t_k-t) / \tau_m} \\
        e^{-(t_k-t) / \tau_m} \\
        \vdots \\
        e^{-(t_k-t) / \tau_m}
    \end{pmatrix}, \quad k = 1, 2, \dots, n
    \label{LIFmemory}
\end{equation}
The exponential function enables efficient memory storage through simple multiplication by $e^{\delta t}$, eliminating complex computations. Equation \eqref{LIFmemory} aligns with state-space models \cite{gu2021efficiently} , connecting state-space models with bio-inspired spiking neural networks as shown in Table \ref{bioinspi}.

\begin{table}[htbp]
\centering
\caption{State-Space Model Perspective of Bio-Inspired Models}
\label{bioinspi}
\begin{tabular}{llll}
\toprule
\textbf{Model} & \textbf{Biological Inspiration} & \textbf{Memory Module} & \textbf{Spiking Mechanism} \\
\midrule
TC-LIF\cite{zhang2024tc} & Multi-compartment & 2D SSM & 1D SSM for spiking \\
PMSN\cite{chen2024pmsn} & Multi-compartment & Multi-dim SSM & IF (parallel trainable) \\
DH-LIF\cite{zheng2024temporal} & Multi-dendrite & Multi-dim SSM & LIF \\
\bottomrule
\end{tabular}
\end{table}
\section{Implementation Details}
\label{sec:pssm}
This section details the implementation of spikingPssm. The model's simplicity allows implementation by adding just a few lines of code to the s4d \cite{gu2021combining} framework,specifically the spiking module in the following algorithm, as spikingPssm essentially combines SSM with PSN\cite{fang2023parallel}'s spiking mechanism.

\begin{algorithm}
\caption{spikingPssm Forward Pass}
\begin{algorithmic}[1]
\State \textbf{Input:} $u$ (shape $B \times H \times L$) \Comment{B: batch size, H: model dimension, L: sequence length}
\State \textbf{Output:} $y$, sparsity ratio $r$
\Statex \rule{\linewidth}{0.4pt}
\State $\textbf{Memory Module}$ 
\State $k \gets \text{S4DKernel}(L)$ 
\State $y \gets \text{FFT\_Conv}(u, k)$
\State $y \gets y +  Du$ \Comment{Core s4d operations}
\Statex \rule{\linewidth}{0.4pt}
\State $\textbf{Spiking Module}$
\State Pad $y$ to length $L + \text{padding}$ 
\State $y1 \gets \text{conv}(y)$ 
\Statex $conv = nn.Conv1d(H, H, kernelsize=reset, stride=reset, bias=False, groups=H)$
\Statex This configuration ensures each neuron only requires one floating-point multiplication per timestep, eliminating cross-channel computations - analogous to LIF's membrane potential decay (multiplied by $1-\frac{1}{\tau}$ each step). Here, $reset$ corresponds to refractory period/convolution kernel length, reducing sequence length to $(L+padding)/reset$.
\State $y1 \gets \text{layernorm}(y1)$
\State $y1 \gets \text{H}(y1)$ \Comment{Heaviside step function}
\State $y\gets y1.\text{repeat\_interleave}(reset)$ \Comment{Replicate each timestep $reset$ times to restore original length}
\Statex \rule{\linewidth}{0.4pt}
\State $\textbf{Nonlinear Channel-Mixer}$
\State $y \gets \text{dropout}(y)$ 
\State $y \gets \text{GLU}(y)$

\State \Return $y$
\end{algorithmic}
\end{algorithm}

\end{document}